\documentclass[10pt,twocolumn,letterpaper]{article}

\usepackage{cvpr}              

\usepackage{graphicx}
\usepackage{amsmath}
\usepackage{amssymb}
\usepackage{booktabs}
\usepackage{verbatim}
\usepackage{subcaption}
\usepackage{float}
\usepackage{amsfonts}
\usepackage{ulem}

\usepackage{ulem} 
\usepackage[ruled,linesnumbered]{algorithm2e}

\usepackage[pagebackref,breaklinks,colorlinks]{hyperref}

\usepackage[capitalize]{cleveref}
\crefname{section}{Sec.}{Secs.}
\Crefname{section}{Section}{Sections}
\Crefname{table}{Table}{Tables}
\crefname{table}{Tab.}{Tabs.}

\def\cvprPaperID{8392} 
\def\confName{CVPR}
\def\confYear{2022}

\begin{document}
\pdfoutput=1
\title{Continual Learning for Pose-Agnostic Object Recognition in 3D Point Clouds}

\author{Xihao Wang\\
Technical University of Munich\\
{\tt\small xihaowang2016@gmail.com}
\and
Xian Wei \\
East China Normal University\\
{\tt\small xian.wei@tum.de} 
\thanks{Corresponding Author}}

\maketitle
\begin{abstract}
Continual Learning aims to learn multiple incoming new tasks continually, and to keep the performance of learned tasks at a consistent level.
%
%
However, existing research on continual learning assumes the pose of the object is pre-defined and well-aligned. For practical application, this work focuses on pose-agnostic continual learning tasks, where the object's pose changes dynamically and unpredictably. 
The point cloud augmentation adopted from past approaches would sharply rise with the task increment in the continual learning process. To address this problem, we inject the equivariance as the additional prior knowledge into the networks. We proposed a novel continual learning model that effectively distillates previous tasks' geometric equivariance information. The experiments show that our method overcomes the challenge of pose-agnostic scenarios in several mainstream point cloud datasets. We further conduct ablation studies to evaluate the validation of each component of our approach. 
\end{abstract}

\section{Introduction}
\label{sec01}
With the development of artificial intelligence, deep neural networks has presented its impressive ability in a wide variety of learning tasks in several fields \cite{dnn1,dnn2}. This outstanding capability partially depends on an ideal learning setting, where sufficient training samples are available and well-aligned. Whereas the world we live in is made up of a series of processes that are changing. With the arrival of the new tasks, the artificial agents are prone to cause serious forgetting of the knowledge learned from old tasks while learning a new task, called catastrophic forgetting \cite{catastrophic}. Thus, continual learning(CL) is introduced to address this problem in order to promote the deep neural networks to satisfy the natural world's requirements \cite{rainbow}.

\begin{figure*}[t]
	\centering
    \includegraphics[width=1.0\textwidth]{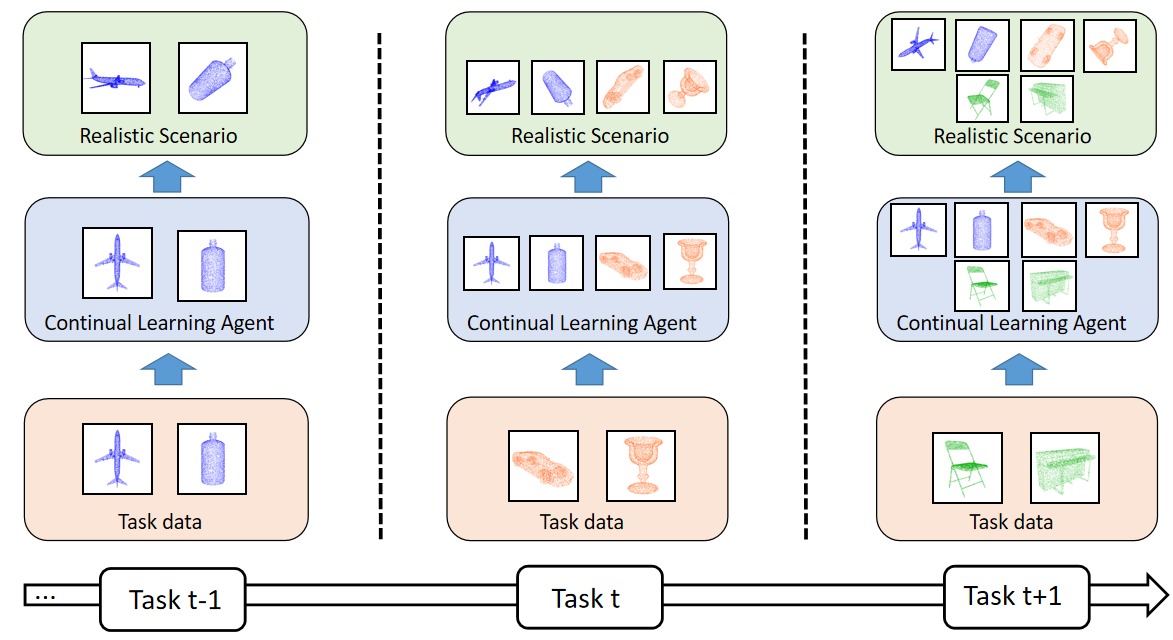}
	\caption{For instance, suppose a robot server that categorizes new items with their 3D point cloud data taken by the surrounding environment. For each category, recognizing new items conspicuously depends on the pose when the robot observes it. The pose of the object varies whenever the robot first learns it or encounters it the next time. We called the above scenario the \textbf{pose-agnostic continual learning}. }
    \label{fig:problem}
\end{figure*}

However, especially in the 3D domain, most of the current continual learning research only evaluated the object in a well-aligned scenario setting. From the view of real-world application, we draw our attention to a more practical scenario of pose-agnostic continual learning. As described in Figure \ref{fig:problem}, the pose-agnostic CL considers the following aspects: (i) The pose of the object is constantly changed in every test phase. (ii) Each task arrives sequentially following the definition of incremental learning. (iii) Due to the limited computing resource, the model could only save a small portion of data from previous tasks into the memory buffers. 
For instance, a robot server is assumed to work in a realistic scenario where it can recognize the object depending on the 3D point cloud data generated from the sensor. 
With the change of surroundings, the robot continually updates the new class captured from its working perception. Due to the constraint of learning resources, the robot could only learn the sample with the first captured pose. In comparison, the object's pose varies every time when the robot encounters the same object. An example is depicted in Figure \ref{fig:problem}, the object bottle learned in task $t-1$ could also be recognized in task $t$ and task $t+1$ even though the pose of the bottle is drastically changed in both tasks.


Despite the fact that the pose-agnostic CL is a practical scenario that is close to our real world, exploration of the 3D geometric data in continual learning is still rare. From the past approaches to 2D data, data augmentation is a prevalent strategy for solving the pose-agnostic problem \cite{augmented}. In recent works of continual learning, data augmentation makes an effort to increase the generalization of given tasks, which not only solves the problem caused by arrived samples' pose but also mitigates the catastrophic forgetting \cite{aug1,aug2}. Nevertheless, the data augmentation strategy is usually difficult to be implemented in the 3D data domain. Due to the increase in freedom degrees, the augmentation of 3D data is always more exhaustive and expensive than 2D data. Furthermore, the computation burden of augmentation would sharply rise with the incremental of the incoming new class in the continual learning process. Thus, we need a strategy to leverage sufficient geometric information from the 3D data and solve the problem caused by data augmentation.

To address this problem, we take our sight on introducing more inductive bias, as prior knowledge, for the pose-agnostic CL. In machine learning, a successful learning scheme usually needs to encode the appropriate  inductive bias \cite{bronstein}. 
For instance, Occam's razor selects a low complexity model \cite{occam}, translational invariance in the convolutional neural network \cite{inductive}, and time invariance in the recurrent neural network\cite{few-shot}.
The popular attention mechanism is also the inductive rule from human intuition \cite{attention}. 
Moreover, the research on the equivariance of the deep neural networks have demonstrated that equivariant features retain more discriminative information  \cite{tfn2,so3spherical,se3transformer,gauge}. To this end, we attempt to design the equivariance as an extra inductive bias into the internal network layers to solve the continual learning problem in the pose-agnostic scenario. 

In this paper, we propose a novel continual learning approach by effectively combining the injected rotation equivariance and distilled knowledge from previous tasks. Firstly, to address the problem of the unpredictable object's pose, we inject the rotation equivariance into the feature extraction block. Then, to cooperate with the equivariant structure, we extend the knowledge distillation framework to leverage the enriched geometric information retained in the network feature maps. Finally, we construct the memory buffer to store the distinguished samples to facilitate the model getting better performance.


Our main contributions are: (i) We develop a dynamic knowledge distillation framework to mitigate catastrophic forgetting, which could efficiently extract equivariant features to overcome the challenge in the pose-agnostic CL scenario.  (ii) Our model presents the effectiveness of exemplars storage and the robustness of encountering complex scenarios. (iii) In the 3D object recognition CL tasks, our method outperforms previous methods in the pose-agnostic scenarios and proposes the corresponding benchmark. 
Furthermore, we conduct ablation studies to prove the validation of each component of our approach. 


\section{Background and Related work}
\label{sec02}
\subsection{Continual Learning} 
\label{sec0201}
The continual learning approaches could be classified into three major categories depending on the storage and usage of task-specific information during the learning process \cite{CLreview}. (i) Replay-based methods mitigate the catastrophic forgetting via replaying the data from past tasks in episodic memories. These methods were inspired by the idea of jointly training the previous samples with the current task \cite{ER,SER,TEM}. Due to the problem of samples imbalance, constrained replay and pseudo rehearsal methods were proposed \cite{DGR,GEM}. 
(ii) Regularization-based methods investigate the continual problem in the no sample conservation situation \cite{EWC,SI,RW,MAS}.  \cite{EWC} constrains the update of network parameters to mitigate forgetting. On the other hand,  \cite{EBLL} solves the continual learning problem by constraining each task's features.
(iii) Dynamic architectural methods leverage the separated network to train different tasks \cite{PackNet,HAT}. The network grows new branches to study current tasks and freezes old branches to remember previous tasks. For other approaches, knowledge distillation which transfers the past version model's knowledge to the next version, could mitigate catastrophic forgetting \cite{LwF,iCaRL}. 
\paragraph{Knowledge Distillation} As one of the approaches in the Dynamic architectural methods, the knowledge distillation transfers the teacher network's knowledge to the student network. The concept of knowledge distillation was firstly proposed in  \cite{KD}, and the works \cite{LwF,iCaRL} also introduced it as one of the parts to solve continual learning problems. Knowledge distillation defines the previous task model as the teacher network. The "dark knowledge" is distilled to the current task model, which is defined as a student network, through the soft labels from the teacher network. Depending on the position of the distillation, the methods could be divided into two categories, distillation from logits \cite{logit1,logit2} and intermediate features \cite{KDin1,KDSD}.

\subsection{Equivariance}
\label{0202}
A function $f:\mathcal{V}\rightarrow\mathcal{U}$ is called equivariant \cite{lieconv} with respect to the transformations $T_{g}:\mathcal{V}\rightarrow\mathcal{V}$ for the abstract group element $g\in G$, if for every $g$ exists a transformation $S_{g}: \mathcal{U}\rightarrow\mathcal{U}$ such that
\begin{equation}
    S_{g}[f(v)]=f(T_{g}[v]),\quad\text{for all}\quad g\in G,v\in \mathcal{V}.
\label{eq:equivariance}
\end{equation}

The $G$ indicates the set of symmetries, and the $g$ is considered as a group element describing the symmetry transformation. If we have two symmetry transformations $(g,g'\in G)$ and we compose them, the result $gg'$ is another symmetry transformation \cite{GroupCNN}. In the Eq. \eqref{eq:equivariance}, the transformation $T_{g}$ and $S_{g}$ are referred to the group action of group element $g\in G$ on object $v\in \mathcal{V}$. Since in the standard point cloud deep learning architectures, like PointNet \cite{pointnet}, PointNet++ \cite{pointnet++}, and DGCNN \cite{dgcnn}, the robustness of rotation is absent. Due to the expensive augmentation in the 3D domain, the equivariance immediately attracts the interest of the researchers. Recent research has demonstrated the importance of the equivariance to ensure stable and predictable performance when the nuisance transformations exist in the data input. \cite{se3transformer}.  
\paragraph{SO(3) Equivariant method} In 3D roto-translation, SO(3) equivariance is a vital property. Based on the SO(3) representation theory, SO(3) equivariant network architecture was proposed. In the early period, spherical convolution leverages the spherical harmonic domain to achieve SO(3) equivariance \cite{sphericalCNN}. Then, a series of works build the equivariance based on steerable kernels. These researches could be divided into two styles, relying on Tensor Field Network theory \cite{TFN,tfn2} or Lie group theory \cite{lieconv,Lietransformer}. Different from explicit designing the steerable kernel,  \cite{VNN} proposed the vector neuron representations for creating SO(3) equivariance implicitly. 

\section{Our Method}
\label{sec03}
\subsection{Problem Definition}
\label{problem_metric}
Among the basic continual learning definition summarized in \cite{threescnario}, we focus on solving the class-incremental problem, which requires the model to infer the task without explicit task identity. We formulate the data arrives incrementally via a batch of point set $x_{i}\in\mathbb{R}^{s_{a}\times d_{v}}\in\mathcal{X}^{i}$ and corresponding true labels $y_{i}\in\mathbb{R}^{b}$, where $s_{a}$ denotes the number of sampled point, $b$ stands for the dimension of the label vector, and $d_{v}$ denotes the dimension of the point depending on whether use normal vector. Each task $\mathcal{T}_{i}$ is composed of point cloud sets and their labels. At each incremental step, point cloud data is only available for new classes $\mathcal{T}_{new}=\{(\mathcal{X}^{c+1},y^{c+1}),\cdots,(\mathcal{X}^{t},y^{t})\}$ and small amount of exemplar data $\mathcal{E}=\{\mathcal{E}^{1},\cdots,\mathcal{E}^{c}\}$. The exemplars are saved from the previous classes $\mathcal{X}_{old}=\{\mathcal{X}^{1},\cdots,\mathcal{X}^{c}\}$. We now formulate the pose-agnostic CL setup. As we mentioned in the introduction, 3D point cloud data has usually suffered the problem of unclear pose-aligned in realistic scenarios. In the pose-agnostic scenario, the input data is presented as $ T_{g}\mathcal{X}^{t}$, where $T_{g}$ denotes the unpredictable SO(3) geometric transformation. 
%
%
As the representation of group element $g\in SO(3)$, $T_{g}$ has a standard rotation matrix that acts on $\mathcal{V}=\mathbb{R}^{3}$.

\label{0302}
\begin{figure*}[h]
	\centering
    \includegraphics[width=1.0\textwidth]{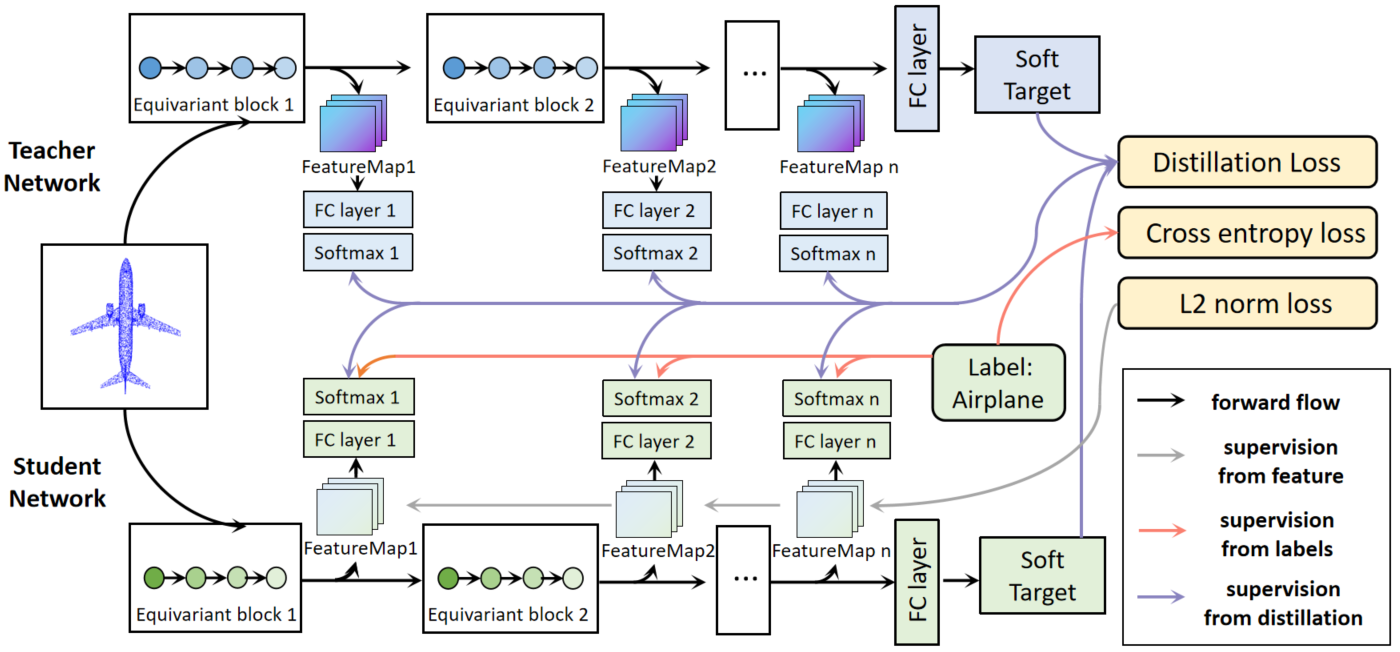}
	\caption{The internal distillation structure consists of multiple loss sources for comprehensive constructing the model's loss function. }
    \label{fig:framework}
\end{figure*}

\subsection{The Basic Learning Framework}
To effectively address the scenario of the pose-agnostic CL, we propose the training model including three components: equivariance injected block, internal distillation structure, and rehearsal exemplar distillation. 
To deal with geometric transformed input point cloud data, we propose to enforce rotation equivariance into the neural networks by designing the feature extractor blocks. The internal distillation structure is responsible for transferring the extracted pose-agnostic representation from the previous tasks to the current tasks. The rehearsal exemplar strategy makes an effort to consolidate the knowledge in the internal distillation structure. 

\subsubsection{Injecting Rotation Equivariance into Network}
The equivariance network is a functional architecture for learning general features with particular inductive bias. In the equivariant network, each layer is required to satisfy the equivariance to transmit the equivariance dependency. Through the above transmission, the semantic information from spatial data is retained between the layers in the network. Thus, the spatial information in each layer is isomorphic with the input point data. In our feature extraction network,  composition blocks could divide the layers into linear, non-linear, and pooling layers. We will describe the SO(3)-equivariance in each layer in the following.

Firstly, the linear layer is the indispensable layer of mapping features in the composition blocks. It could be represented as $\mathcal{U}=f(\mathcal{V})=W\mathcal{V}$. When the rotated input passes through the linear layer, the representation could be written as:  
\begin{equation}
    f(T_{g}\mathcal{V})=W\mathcal{V}T_{g}=T_{g}f(\mathcal{V})=T_{g}W\mathcal{V},
\label{eq:linear}
\end{equation}
which indicates that the linear layer naturally possesses the SO(3)-equivariance property. Thus, the linear layer could maintain its original construction. 


Secondly, the non-linear layer usually makes a critical effort on the neural networks' learning ability. Inspired by the work \cite{VNN}, we also applied the implicit representation to predict the direction of the input vector feature dynamically. At the non-linear layer, the input vector is the output of the preceding linear layer, $q=f_{lin}(\mathcal{V})=W\mathcal{V}$. Moreover, $k=\Phi\mathcal{V}$ is the normal direction of $q$. Depending on the distance of inner product space, the output of the non-linear layer $\mathcal{U}=f(\mathcal{V})=\mathcal{V}$, when $\langle q, k\rangle \ge 0$. Otherwise, $\mathcal{U}=f(\mathcal{V})=q-\left\langle q, \frac{k}{\|k\|}\right\rangle \frac{k}{\|k\|}$. Since the rotation transformation does not influence the distance in the inner product space, the SO(3)-equivariance could be verify as 
\begin{equation}
\begin{aligned}
    f(T_{g}\mathcal{V})=T_{g}q-\left\langle T_{g}q, \frac{T_{g}k}{\|T_{g}k\|}\right\rangle \frac{T_{g}k}{\|T_{g}k\|} \\
    =T_{g}(q-\left\langle q, \frac{k}{\|k\|}\right\rangle \frac{k}{\|k\|})=T_{g}f(\mathcal{V}), \text{if}\langle q, k\rangle \le 0.
\end{aligned}
\label{eq:non-linear}
\end{equation}

Lastly, according to the blueprint of \cite{bronstein}, the pooling, which makes the function of aggregating the feature, could be divided into two types. In the case of global pooling, it is the invariance layer. About the case of local pooling, we focus on the maximum pooling, where the SO(3)-equivariant construction follows the insight from the non-linear layer \cite{VNN}.
%
%
Through the learned direction $\xi=\Psi\mathcal{V}$, the maximum pooling chooses the input vector element that best aligns with the learned direction $d^{*}$. The SO(3)-equivariance could be verified as follows: 
\begin{equation}
    f_{max}(T_{g}\mathcal{V})=\underset{d^{*}}{\operatorname{argmax}}\langle T_{g}\xi,T_{g}\mathcal{V}\rangle =T_{g}f_{max}(\mathcal{V}).
\label{eq:maxpooling}
\end{equation}


\subsubsection{Internal Distillation Structure}
After the SO(3)-equivariance is injected into the feature extraction network, the model requires a structure to resist the catastrophic forgetting by transferring learned knowledge from the current model $\Theta^{c}$ to the target model $\Theta^{t}$. Even though more than one kind of method could achieve continual learning, we choose knowledge distillation as our approach because it could effectively transfer the equivariance learned by the previous model version \cite{few-shot}. However, previous knowledge distillation works did not adapt to our equivariant feature extraction network $\zeta(\cdot)$. 
%
Due to the reason that our feature extraction network retains the rich semantic information in the feature map, we would not adopt the typical knowledge distillation framework that only utilizes the logits of the output layer. We leverage the feature map between sections to combine with the internal distillation block. Thus, the knowledge retained in the network is squeezed into the shallow portion to affect the final full-connect layer (illustrated in Figure \ref{fig:framework}) 

During the training process, the selected feature between sections is mapped to an external layer individually to keep a unified dimension. Similar to the typical knowledge distillation, the information in the external layer is distilled and conveyed to constitute a comprehensive final loss as:
\begin{equation}
\text{loss}_{\mathcal{X}}=\sum^{D}_{\delta=1}(loss_{1}+loss_{2}+loss_{3}+loss_{4}).
\end{equation}
The final loss has four sources. To construct the loss about hard labels in the student's model, the first source is the cross-entropy from the ground truth and each feature map in the target model:
\begin{equation}
loss_{1}=-(1-\gamma)\sum^{t}_{i=c}y^{i}\cdot\log(\sigma(\Theta^{t}(\mathcal{X}^{i}_{\delta}))),
\label{eq:source1}
\end{equation}
where $(\mathcal{X},y)_{i=c}^{t}$ come from the input samples in the new task, and $\Theta(\mathcal{X}_{\delta})$ denotes the logits from the feature map. The first source not only employs the final classifier layer but also allows every feature map, which contains the equivariance semantic information from the input, participate in the computation of the cross-entropy. Then, in order to compute the distillation loss between soft labels and soft predictions, the second is the KL-divergence between the student model's feature map and the corresponding feature map trained in the teacher model:
\begin{equation}
    loss_{2}=\lambda\text{KL}(\sigma(\Theta^{t}(\mathcal{X}_{\delta})/T), \sigma(\Theta^{c}(\mathcal{X}_{\delta})/T)).
\label{eq:source2}
\end{equation}
Here, the second source connects the distribution of each feature map from the teacher's model to the student's model. Next, the third source is the KL-divergence between each internal feature map and the final output:
\begin{equation}
    loss_{3}=\gamma\text{KL}(\sigma(\Theta^{t}(\mathcal{X}_{\delta})/T), \sigma(\Theta^{t}(\mathcal{X}_{D})/T)),
\label{eq:source3}
\end{equation}
where $\Theta^{t}(\mathcal{X}_{D})$denotes the final classifier logits. The third source makes the feature map which hides in the deep layer to affect the final classifier output in the most shallow layer. Lastly, in the aspect of the Euclidean space, the fourth source enforces the "dark knowledge" in each feature map to the shallow output layer:
\begin{equation}
    loss_{4}=\kappa||F_{\delta}-F_{D}||^{2}_{2},
\label{eq:source4}
\end{equation}
where $F_{\cdot}$ denotes the feature map of indicated layer. Depending on the above comprehensive loss constitution, the rich semantic information extracted by the equivariance network sufficiently facilitates the student network to inherit the knowledge from the teacher network. 

\subsubsection{Rehearsal Exemplar Distillation}
According to the outstanding robustness and performance of the rehearsal, we also employ the memory buffer to save the selected exemplar of old classes.
Thus, in each iteration, the input of trained model contains the current task's data $\mathcal{X}_{new}$ and the storage exemplar $\mathcal{E}_{new}$. Since the redundancy input augmentation has been avoided by enforcing the SO(3)-equivariance into the network, the burden of saving exemplar sets of the new tasks has been remarkably relieved. The distillation of rehearsal exemplar consists of two steps: exemplar selection and exemplar distillation. 

\paragraph{Exemplar Selection}
Among the input data $\mathcal{X}_{new}$, we will randomly select a certain number of samples to supply into the $\mathcal{E}_{old}=\mathcal{E}$ for creating new exemplar set $\mathcal{E}_{new}=\{\mathcal{E}^{1},\cdots,\mathcal{E}^{t}\}$. We define that the rehearsal memory buffer could save up to $M$ samples. Meanwhile, in order to keep a balance of the number of saved samples, we randomly selected an equal number of samples for each class when they arrived with the new task. Hence, the sample number of each class is denoted as $r=M/t$ in the updated exemplar $\mathcal{E}_{new}$. With the class increase, it is necessary to discard some samples from each class to keep the total number of samples in the exemplar. We assume that the center of the $\mathcal{E}^{i}$ is the mean of the feature vector, and the $r$ nearest samples to the center would be picked in Euclidean distance.
\begin{equation}
    d_{k}=||\frac{1}{r}\sum^{r}_{i=1}\zeta(e_{i})-\zeta(e_{k})||^{2}_{2},\quad \text{for all}\quad e_{k}=\{e_{1},\cdots,e_{r}\}.
\label{eq:exemplar_selection}
\end{equation}
Depending on the distance, the last $r'-r$ samples will be discarded. 

\paragraph{Exemplar Distillation}
In order to balance the effect between new and old exemplar samples, the source of the exemplar distillation consists of two elements. 
\begin{equation}
\begin{aligned}
    \text{loss}_{\mathcal{E}}=-\sum^{c}_{i=1}\hat{y}^{i}\cdot\log(\sigma(\Theta^{t}(\hat{\mathcal{X}^{i}}))) \\ +\lambda\text{KL}(\sigma(\Theta^{t}(\hat{\mathcal{X}})/T), \sigma(\Theta^{c}(\hat{\mathcal{X}})/T)),
\end{aligned}
\label{eq:exemplar_distillation}
\end{equation}
where $\hat{\mathcal{X}}, \hat{y}$ denotes the sample from exemplar $\mathcal{E}_{new}$, $\sigma(\cdot)$ indicates the softmax function, $T$ stands for the temperature of distillation, and $\lambda$ is the hyper-parameter. The first element is the cross-entropy of the exemplar samples and their labels. The second element is the KL(Kullback-Leibler) divergence loss to minimize the Euclidean distance between the corresponding exemplar logits. The completed process is presented in Algorithm\ref{algorithm}.  

\begin{algorithm}[ht]
    \caption{Pose-agnostic Continual Learning Algorithm}\label{algorithm}
\SetKwInOut{Input}{Input}\SetKwInOut{Output}{Output}

\Input{new task data $\mathcal{T}_{new}=\{(\mathcal{X}^{c+1},y^{c+1}),\cdots,(\mathcal{X}^{t},y^{t})\}$, old exemplar set $\mathcal{E}_{old}=\{\mathcal{E}^{1},\cdots,\mathcal{E}^{c}\}$, current model $\Theta^{c}$, distillation temperature $T$, Memory size $M$}
\Output{target model trained up to t classes $\Theta^{t}$, new exemplar set $\mathcal{E}_{new}$}
update the memory size $r=M/t$\;
\For{$(\mathcal{X}^{i},y^{i})\in\mathcal{T}_{new}$}
{Randomly pick $r$ samples $\mathcal{E}^{i}\rightarrow \{e_{1},\cdots,e_{r}\}\subset\mathcal{X}^{i}$ \; 
Softmax over the logtis from each internal feature map in target model $\sigma(\Theta^{t}(\mathcal{X}^{i}_{\delta}))$\;
Compute the classification loss Eq. \eqref{eq:source1} and distillation loss Eq. \eqref{eq:source3} in target model\;
Compute the Feature map loss Eq. \eqref{eq:source4} in L2 norm\;
Softmax over the logtis from each internal feature map in current model $\sigma(\Theta^{c}(\mathcal{X}^{i}_{\delta}))$\;
Compute the distillation loss Eq. \eqref{eq:source2} between current model and target model\;
\textbf{New exemplar set selection}\;
Randomly pick $r$ samples $\mathcal{E}^{i}\rightarrow \{e_{1},\cdots,e_{r}\}\subset\mathcal{X}^{i}$ \; 
Calculate the mean feature $\xi(e_{i})$\;
Arrange the $\mathcal{E}^{i}\subset\mathcal{X}^{i}$ in descending order depending on the distance in Eq. \eqref{eq:exemplar_selection}\;
\textbf{Load the input exemplar data}\;
\For{$(\hat{\mathcal{X}^{i}},\hat{y}^{i})\in\mathcal{E}_{old}$}
{Sample number of each class in $\mathcal{E}_{old}\rightarrow r'$\;
Compute the distillation loss of exemplar set Eq. \eqref{eq:exemplar_distillation}\;
Discard last $r'-r$ old samples to generate $\hat{\mathcal{E}^{i}}\subset\mathcal{E}_{old}$\;
}
$\text{Loss}=\text{loss}_{\mathcal{X}}+\text{loss}_{\mathcal{E}}$\;
$\mathcal{E}_{new}\leftarrow \{\hat{\mathcal{E}^{1}},\cdots,\hat{\mathcal{E}^{c}},\mathcal{E}^{i}\}$\;
}
\end{algorithm}

\section{Experiment}
\label{sec:experiment}
In this section, we investigate the capabilities of our method and other continual learning approaches to two popular 3D point cloud datasets(ModelNet40 \cite{ModelNet40} and ScanObjectNN \cite{ScanObjectNN}) for the object recognition task and achieve consistent improvement. Especially in the pose-agnostic case, our model overcomes the challenge of unpredictable rotation, which is usually encountered in realistic scenarios. We describe the dataset and implementation details in the section \ref{sec:dataset}. Following the results, the comparative analysis and ablation studies are presented. 
\begin{table*}[t]
\centering
{
\begin{tabular}{|c|cccccccccc|c|}
\hline
Methods              & 4   & 8     & 12    & 16    & 20    & 24    & 28    & 32    & 36    & 40    & Avg   \\ 
\hline
\multicolumn{12}{|c|}{Aligned/Aligned}\\
\hline
LwF \cite{LwF} & 96.5  & 87.2  & 77.5  & 70.6  & 62.3  & 56.8  & 44.7  & 39.4  & 36.1  & 31.5  & 60.3  \\ 
iCaRL \cite{iCaRL} & 96.8  & 90.4  & 83.6  & 78.3  & 72.5  & 67.3  & 59.6  & 53.1  & 47.8  & 39.6  & 68.9  \\ 
DeeSIL \cite{DeeSIL}& 97.7  & 91.5  & 85.4  & 80.5  & 74.4  & 71.8  & 65.3  & 58.7  & 52.4  & 43.7  & 72.1  \\
EEIL \cite{EEIL} & 97.6  & 93.8  & 87.5  & 81.6  & 78.2  & 74.7  & 69.2  & 62.4  & 56.8  & 48.1  & 75.0    \\ 
IL2M \cite{IL2M}& 97.8  & \textbf{95.1}  & 89.4  & 85.7  & 83.8  & 82.2  & 78.4  & 72.8  & 67.9  & 57.6  & 81.1  \\ 
DGMw \cite{DGM}& 97.5  & 93.2  & 86.4  & 82.5  & 80.1  & 78.4  & 73.6  & 65.3  & 61.5  & 53.4  & 77.2  \\ 
DGMa \cite{DGM}& 97.5  & 93.4  & 84.7  & 81.8  & 79.5  & 77.8  & 74.1  & 67.4  & 60.8  & 51.5  & 76.8  \\ 
BiC \cite{BiC} & 97.8  & 95.5  & 88.5  & 86.9  & 84.3  & \textbf{83.1}  & \textbf{79.3}  & 74.2  & 70.7  & 59.2  & \textbf{82.0}    \\ 
RPS-Net \cite{RPS-Net}& 97.7  & 94.6  & \textbf{90.3}  & \textbf{88.2}  & \textbf{86.7}  & 82.5  & 78.0    & 73.6  & 68.4  & 58.3  & 81.7  \\
\textbf{Ours} & \textbf{99.6} & 92.8 & 88.5 & 87.3 & 85.3 & 81.9 & 78.7 & \textbf{74.8} & \textbf{71.8} & \textbf{61.2} & \textbf{82.0} \\
\hline
\multicolumn{12}{|c|}{Pose-Agnostic/Pose-Agnostic}\\
\hline
iCaRL & 98.7  & 61.8  & 37.7  & 32.2  & 28.8  & 27.9  & 24.6  & 19.2  & 15.6  & 14.1  & 36.1  \\ 
iCaRL with EQ & 98.3 & 79.1 & 72.5 & 71.0 & 69.8 & 66.9 & 64.2 & 58.8 & 52.4 & 44.2 & 67.7\\
Ours w/o EQ & 99.2 & 79.2 & 64.8 & 54.6 & 45.8 & 39.2 & 36.3 & 26.1 & 21.8 & 17.6 & 48.5\\
Ours w/o EM & 98.3 & 86.3 & 65.4 & 51.7 & 47.1 & 40.8 & 37.6 & 34.1 & 26.4 & 25.8 & 51.4\\
\textbf{Ours} & 99.6 & 94.8 & 88.5 & 88.3 & 87.3 & 81.9 & 78.7 & 76.8 & 71.8 & 63.2 & \textbf{83.2} \\ \hline
\end{tabular}
}
\caption{Quantitative comparisons on ModelNet40 dataset. iCaRL-EQ denotes the iCaRL method injected SO(3)-equivariance. Ours-no EM indicates our model only without ememplar memory. Ours-no EQ denotes our model only removes SO(3)-equivairance.}
\label{tab:exp_modelnet}
\end{table*}

\begin{figure*}[ht]
	\centering
    \includegraphics[width=1.0\textwidth]{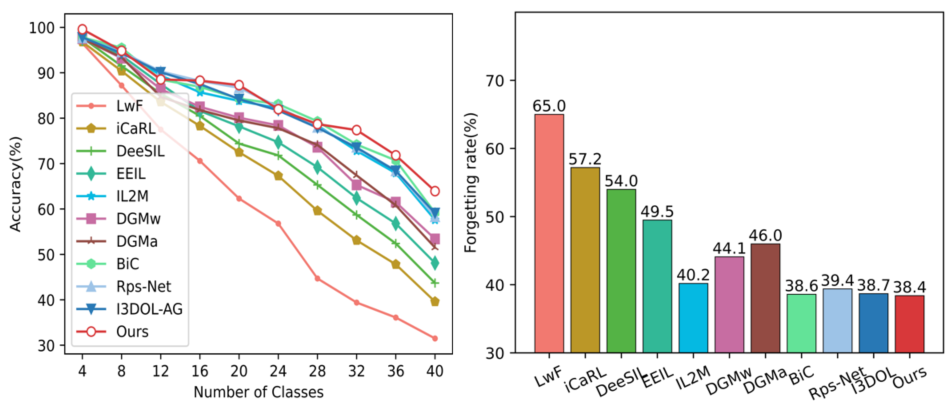}
	\caption{Effect Investigation about Accuracy Tendency and Forgetting Rate on ModelNet40 dataset }
    \label{fig:modelnet}
\end{figure*}
\subsection{Datasets and Implementation details}
\label{sec:dataset}
According to the datasets employed, the ModelNet40 \cite{ModelNet40}, which consists of clean 3D CAD object models from 40 classes, is divided into 9843 training samples and 2468 testing samples. We define the number of continual learning tasks as 10, and each task $\mathcal{T}_{i}$ has 4 classes model. About the size of the exemplar memory, $M$ is 500. For the ScanObject dataset \cite{ScanObjectNN},  it contains 15000 point cloud objects which are collected from 2902 real-world unique object instances into 15 categories. The number of continual learning tasks is 5, and each task $\mathcal{T}_{i}$ has 3 classes model.The memory size $M$ for the ScanObjectNN dataset is 400. We defined the training and test environment to simulate the pose-agostic case in realistic scenarios, where the $\cdot/\cdot$ denotes the train/test situation. The \textbf{Aligned} denotes the pose of the model is well-aligned, and the \textbf{Pose-Agnostic} denotes the pose of the model is unknown due to the arbitrary rotation. We train our model use a server with
8 GeForce RTX 3090(24GB).

\subsection{Evaluation Metrics} 
We employ three metrics to evaluate the performance of our CL model: average accuracy, forgetting rate, and feature retention. In the test phase, the test samples are also possessed agnostic geometric transformation to simulate the real world. (i) Average accuracy is represented as \textbf{(Avg Acc)}, which is similar to  \cite{iCaRL}.(ii) Forgetting rate $\mathcal{F}$ measures the accuracy drop from the first task, as the definition of  \cite{forgetrate}. (iii) Following the  \cite{ECIL}, the feature retention $\mathcal{R}$ measures how much information is retained in the feature extractor. After the entire learning process is finished, the $\mathcal{R}$ could be computed as te performance ratio between the final model and the model of the current task is arrived.

\subsection{Evaluation in the ModelNet40 dataset}
As described in Table\ref{tab:exp_modelnet}, we conduct the object classification task in ModelNet40 dataset. From the results of per-task accuracy, our model notably surpasses the other approaches on the average accuracy metric in the case of Aligned/Aligned. In the case of pose-agnostic/pose-agnostic, our model overcomes the challenge of maintaining satisfactory performance and avoiding catastrophic forgetting in an unfamiliar environment where the pose of the object is changeable. Our model achieves the consistently the same result in both cases, but other methods attain a poor performance compared with their result in the well-aligned case. 

The figure\ref{fig:modelnet} illustrates the change tendency of the accuracy when a new task arrives. Our model keeps a relatively high result in each task and gets the best average accuracy \textbf{Avg Acc}. On the other hand, our model presents the lowest forgetting rate $\mathcal{F}$ compared with other methods. The ratio of information retained $\mathcal{R}$ could be presented as the slope of the accuracy. In the figure\ref{fig:modelnet}, our model exhibits a smooth from the first task to the final task, which stands for the knowledge from the last model is transferred to the next model sufficiently. 
\begin{figure*}[h]
	\centering
    \includegraphics[width=1.0\textwidth]{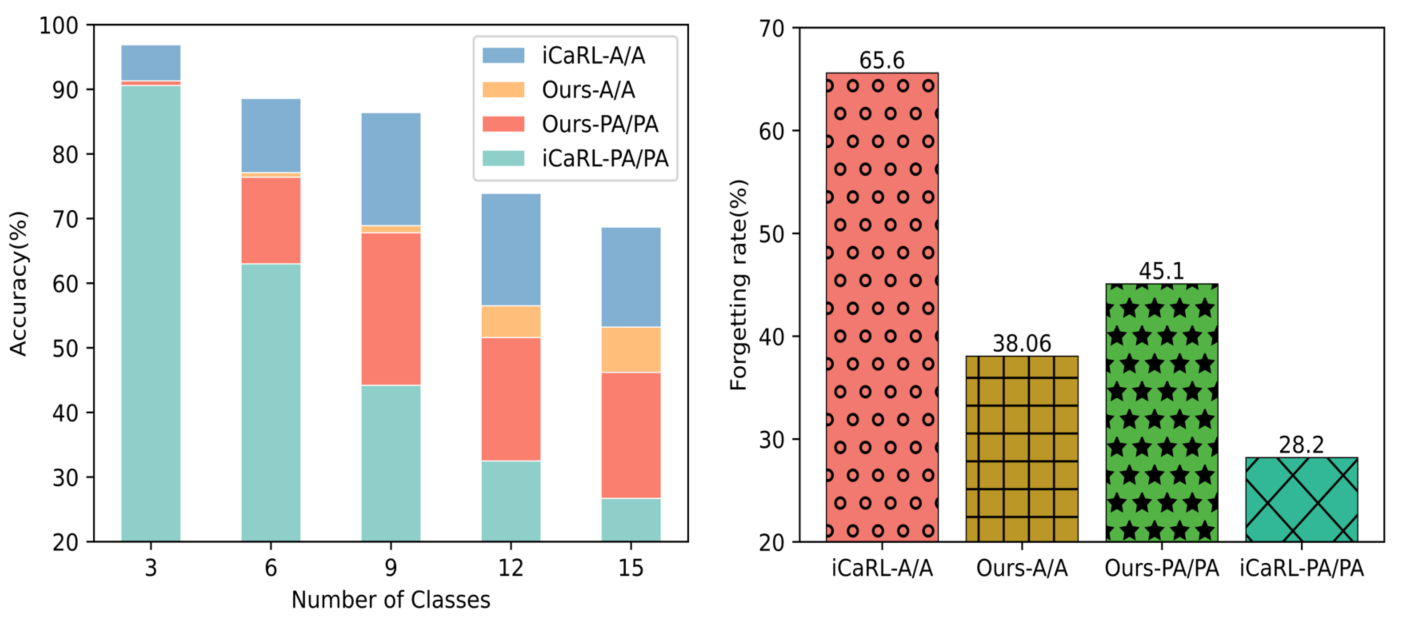}
	\caption{Effect Investigation about Accuracy Tendency and Forgetting Rate on ScanObjectNN dataset. A/A denotes Aligned/Aligned situation. PA/PA indicates Pose-Agnostic/Pose-Agnostic situation }
    \label{fig:shapenet}
\end{figure*}

\paragraph{Ablation Studies}
We conduct ablation studies to explore the effectiveness of different components in our model in Table \ref{tab:exp_modelnet}. One of the studies investigates the performance of equivariance injection. We additional enforced the equivariance into the iCaRL method, but it did not reach the performance where the object pose is well-aligned. Although injecting equivariance into the network could alleviate the difficulty of the pose-agnostic case. It still requires a suitable structure to support the achievement of good performance. On the other hand, we also conducted the test to remove the equivariance of our model. Even though the original architecture got a high result in the first task, the (\textbf{Acc}) decreased significantly with the new task arriving. The result proves that the equivariance makes a significant effort to our model. Moreover, the exemplar memory also has a high contribution to the continual learning task. Thus, the study proves that each component of our model plays an indispensable role in achieving desired performance.

\subsection{Evaluation in the ScanObjectNN dataset}
\label{sec:ScanObjectNN}
To challenge a more realistic scenario, we conducted the experiment on the ScanObjectNN dataset to evaluate our model. Different from the ModelNet40 dataset, the ScanObjectNN dataset is obtained based on scanned indoor scene data with background noise. From the Table\ref{fig:shapenet}, we can note that our model also defeats the challenge of the pose-agnostic problem. In the case of the iCaRL method, it performs well in the well-aligned scenario. However, its average accuracy gets a marked drop under the pose-agnostic scenario. In contrast, our method performs similarly between the two different scenarios. Due to the more complex environment, our model also exposed its limitation that our model does not outperform the performance of the iCaRL method in the well-aligned scenario. But we believe that our performance still has improvement space with the further fine-tuning process.


\section{Conclusion}
In this paper, we address a realistic class incremental continual learning scenarios where the pose of the object is dynamically changed. To effectively address such a scenario, we propose to inject the rotation equivariance as the additional prior knowledge into the network and design a novel training framework to alleviate the catastrophic forgetting. Our model achieves competitive performance in both well-aligned and pose-agnostic scenarios by injecting the SO(3)-equivariance into the network. The experiment results on popular point cloud datasets demonstrate the effectiveness of our model. Specifically, our model sufficiently leverages the information among feature maps in the network to improve the efficiency of knowledge distillation. In addition, our approach releases the redundancy data augmentation in the 3D domain. In future work, we will further investigate the relationship between other prior knowledge and continual learning performance. We believe that our model will facilitate continual learning approaches to address real-world problems.


{\small

\bibliographystyle{ieee_fullname}
\normalem
\bibliography{egbib}
}


\end{document}